\documentclass{article}

\PassOptionsToPackage{numbers}{natbib}

\usepackage[preprint]{neurips_2024}

\usepackage[utf8]{inputenc} 
\usepackage[T1]{fontenc}    
\usepackage{url}            
\usepackage{booktabs}       
\usepackage{amsfonts}       
\usepackage{nicefrac}       
\usepackage{microtype}      
\usepackage{xcolor}         
\usepackage{amsmath}
\usepackage{bm}
\usepackage{graphicx}
\usepackage{tikz}
\usepackage{multirow}
\usepackage{colortbl}
\usepackage{wrapfig}
\usepackage{caption2}

\definecolor{codeblue}{rgb}{0.25,0.5,0.5}
\definecolor{myblue}{rgb}{0.88,0.98,1}
\definecolor{mygreen}{rgb}{0.92, 1.0, 0.92}
\definecolor{myred}{rgb}{1, 0.9, 0.9}
\definecolor{mygray}{gray}{0.95}

\definecolor{Highlight}{HTML}{E8F8F5}
\definecolor{midgreen}{HTML}{589d62}
\definecolor{midblue}{HTML}{69a3f1}
\definecolor{darkgreen}{HTML}{146038}
\definecolor{darkblue}{HTML}{143b59}
\definecolor{hotpink}{RGB}{59, 115, 227}

\usepackage[pagebackref=true,breaklinks=true,colorlinks,bookmarks=false,citecolor=hotpink]{hyperref}

\title{Benchmarking and Improving Detail Image Caption}

\author{
Hongyuan Dong\textsuperscript{1}\textsuperscript{*}, 
Jiawen Li\textsuperscript{1}\textsuperscript{*}, 
Bohong Wu\textsuperscript{1}, 
Jiacong Wang\textsuperscript{1,2}, 
Yuan Zhang\textsuperscript{1,3}, 
Haoyuan Guo\textsuperscript{1}\textsuperscript{\dag}\\
\textsuperscript{1}ByteDance Inc. \quad \textsuperscript{2}School of Artificial Intelligence, University of Chinese Academy of Sciences \\
\textsuperscript{3}School of Computer Science, Peking University \\
\texttt{\{donghongyuan.dousia, lijiawen.0818, bohongwu\}@bytedance.com} \\
\texttt{wangjiacong20@mails.ucas.ac.cn, \{zhangyuan.gump, guohaoyuan\}@bytedance.com} \\
}

\begin{document}

\maketitle
\renewcommand{\thefootnote}{}
\footnotetext{* Equal contribution.}
\footnotetext{\dag Email corresponding}

\begin{abstract}
Image captioning has long been regarded as a fundamental task in visual understanding. 
Recently, however, few large vision-language model (LVLM) research discusses model's image captioning performance because of the outdated short-caption benchmarks and unreliable evaluation metrics. 
In this work, we propose to benchmark detail image caption task by curating high-quality evaluation datasets annotated by human experts, GPT-4V, Gemini-1.5-Pro and GPT-4O.
We also design a more reliable caption evaluation metric called \textbf{CAPTURE} (CAPtion evaluation by exTracting and coUpling coRE information).
CAPTURE extracts visual elements, e.g., objects, attributes and relations from captions, and then matches these elements through three stages, achieving the highest consistency with expert judgements over other rule-based or model-based caption metrics. 
The proposed benchmark and metric provide reliable evaluation for LVLM's detailed image captioning ability. 
Guided by this evaluation, we further explore to unleash LVLM's detail caption capabilities by synthesizing high-quality data through a five-stage data construction pipeline. 
Our pipeline only uses a given LVLM itself and other open-source tools, without any human or GPT-4V annotation in the loop.
Experiments show that the proposed data construction strategy significantly improves model-generated detail caption data quality for LVLMs with leading performance, and the data quality can be further improved in a self-looping paradigm.
All code and dataset will be publicly available at \href{https://github.com/foundation-multimodal-models/CAPTURE}{https://github.com/foundation-multimodal-models/CAPTURE}.

\end{abstract}

\section{Introduction}
\label{sec: intro}

Image captioning has long been a fundamental task to assess LVLM's vision understanding capability~\cite{Wang2023CogVLMVE,liu2024llavanext,chen2023internvl,dong2024internlm}.
However, recent LVLM researches evaluate LVLMs' visual understanding performance with a focus on QA benchmarks, such as MME~\citep{Fu2023MMEAC}, MMBench~\citep{Liu2023MMBenchIY}, MMMU~\citep{yue2023mmmu}, MM-Vet~\citep{Yu2023MMVetEL}, etc., which may suffer from instability caused by LVLMs' varying instruction following abilities~\citep{Fu2023MMEAC}. 
What's worse, human-defined queries may cover a limited scope of vision features~\citep{Li2023SEEDBenchBM} and introduce bias in performance evaluation~\citep{Yu2023MMVetEL}.
Traditional image captioning task is considered unreliable for visual understanding evaluation because of the outdated benchmarks and unstable evaluation metrics. 
Current image caption benchmarks consist of fairly short captions with limited vision information~\citep{lin2014microsoft, agrawal2019nocaps}, while SOTA LVLMs are capable of generating detail image captions encompassing a variety of fine-grained elements~\citep{chen2023sharegpt4v, Wang2023CogVLMVE, liu2024llavanext}, and only a few of them are covered in the provided ground truth captions.
This contradiction leads to unsatisfying evaluation results.  
To this end, we propose to curate high-quality detail image caption evaluation datasets to provide reliable evaluation results for SOTA LVLMs. 
The evaluation datasets are annotated by human experts and the most capable LVLM GPT-4V\cite{gpt4v}, Gemini-1.5-Pro\cite{reid2024gemini} and GPT-4O~\cite{gpt4o}, and are therefore of satisfying quality for state-of-the-art (SOTA) LVLM evaluation. 

Apart from benchmarks, existing caption evaluation metrics also suffer from poor consistency with human judgements. 
Traditional rule-based caption metric such as BLEU~\citep{papineni2002bleu}, CIDER~\citep{vedantam2015cider} and METEOR~\citep{banerjee2005meteor}, 
compute \textit{n}-gram segment matching score between candidate and reference captions, which is extremely sensitive to caption writing style, resulting into unstable evaluation results~\citep{hessel-etal-2021-clipscore}. 
Model-based evaluation metric are proposed to improve the reliability of image caption evaluation. 
However, representative model-based metrics either adopt outdated backbone models~\cite{anderson2016spice}, or suffer from limited input text length~\cite{hessel-etal-2021-clipscore, sarto2023positive}, leading to unsatisfying detail caption evaluation results.

\begin{figure*}[t]
\centering
\begin{tikzpicture}
\draw (0,0 ) node[inner sep=0] {\includegraphics[width=\columnwidth, trim={0.2cm 1cm 0.2cm 0.2cm}, clip]{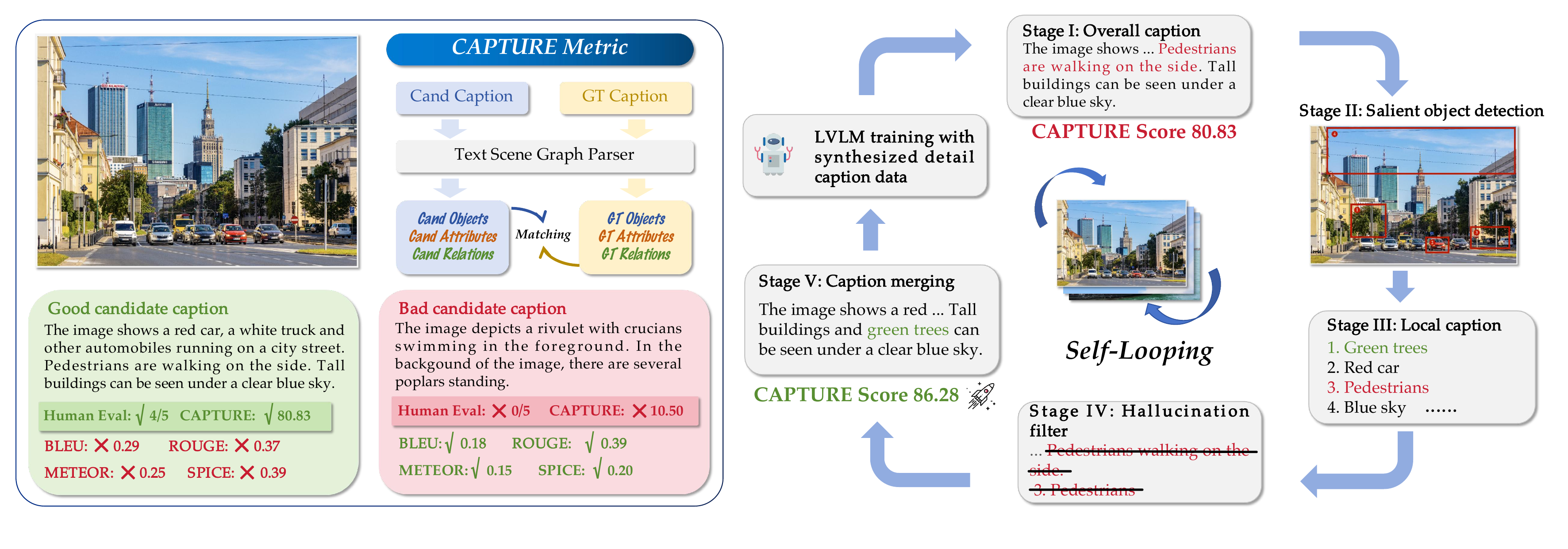}};
\end{tikzpicture}
\caption{
An illustration of the proposed detail caption evaluation metric CAPTURE and caption data quality improvement pipeline. 
}\label{fig: intro}
\vspace{-3mm}
\end{figure*}

To tackle the aforementioned problems, we propose CAPTURE, which adopts the SOTA text scene graph parser Factual~\citep{li2023factual} to extract visual elements from captions, i.e., objects, attributes and relations. 
We match the extracted elements from candidate and ground truth captions through a stop words filtering module and a three-stage matching strategy.
Compared with SPICE, our proposed CAPTURE metric adopts a T5-based language model as parser rather than PCFG, while we design a more capable three-stage core information coupling module to match the parsed result.
As illustrated in Figure~\ref{fig: intro}, CAPTURE produces satisfying consistency with human evaluation results, while other metrics do not.
Experiments on both GPT-4 annotated dataset and human-annotated datasets show that the proposed CAPTURE achieves the highest consistency with human or GPT-4 experts, surpassing all traditional caption evaluation metrics and model-based metrics.

With CAPTURE providing reliable evaluation results, we further explore to unleash LVLMs' detail image caption capabilities in a divide-and-conquer paradigm with a given LVLM.
No expert annotation is required in our proposed data construction loop.
The data construction pipeline is illustrated in Figure~\ref{fig: intro}. 
We adopt a divide-and-conquer strategy to synthesize high-quality detail image caption.
An LVLM is instructed to generation both overall caption for the image and local captions for salient objects segmented by SAM~\cite{kirillov2023segment}.
We adopt a novel phrase-level filtering strategy to suppress hallucinations, which extracts visual element phrases from captions, and filter out those scored low by the open-vocabulary object detection model. 
Finally, the filtered overall caption and local captions are fed to an LLM to be merged into a high-quality detail image caption. 
Experiments show that our data construction pipeline produces significantly higher-quality detail caption, and a simple-yet-effective self-looping strategy can further improve the data quality. 
Moreover, the synthesized data improves LVLM's understanding capabilities effectively when incorporated into the training process.


To summarize, the contribution of this work can be listed as follows:

\textbf{(1)} We release a 4870-case GPT-4V, Gemini-1.5-Pro and GPT-4O annotated detail image caption benchmark for reliable evaluation, accompanied with three model-generated captions and corresponding GPT-4 annotated quality scores for expert judgement consistency evaluation. 

\textbf{(2)} We propose a novel detail image caption evaluation metric CAPTURE, which adopts a T5-based parser to extract visual elements from captions, and compute the matching score via a three-stage matching module. 
Experiments indicate that CAPTURE metric achieves the highest consistency with expert judgement over other caption metrics, providing reliable detail caption evaluation results without expensive LLM API calls. 

\textbf{(3)} We propose a five-stage detail image caption data construction pipeline, which explores to use a given LVLM and open-source vision and language tools to produce higher-quality detail caption data.
Experiments show that our data construction pipeline improves detail caption data quality significantly, and the data quality can be further improved by self-looping. 


\section{Related Work}
\label{sec: bg}
\paragraph{Image caption evaluation. }
Early image captioning benchmarks, such as COCO~\cite{chen2015microsoft}, NoCaps~\cite{agrawal2019nocaps}, consist of precise annotated captions but contain limited visual information, which is outdated for recently released LVLMs with leading performance. 
Traditional caption evaluation metrics, such BLEU~\citep{papineni2002bleu}, CIDER~\citep{vedantam2015cider} and METEOR~\citep{banerjee2005meteor}, compute \textit{n}-gram matching score and therefore suffer from instability caused by varying writing styles.  
Model-based metric SPICE~\citep{anderson2016spice} extracts visual elements from caption sentences, and match them to obtain evaluation results.  
CLIP-Score~\citep{hessel-etal-2021-clipscore}, MID~\citep{kim2022mutual} and PAC-S~\citep{sarto2023positive} borrow pretrained CLIP~\citep{radford2021learning} model to assess the quality of model-generated image captions. 
Although producing relatively reliable evaluation results, these metrics can hardly tackle detail caption evaluation tasks because of the outdated backbone model (SPICE) and limited text input length (CLIPScore). 

\paragraph{Detail caption data construction. }
A series of work seek to construct detail caption data for LVLM training. 
ShareGPT4V~\cite{chen2023sharegpt4v} and ALLaVA~\cite{chen2024allava} curate detail image caption data annotated by GPT-4V for model training. 
All-Seeing~\cite{wang2023all} leverages LLMs to imagine co-occurrence visual elements for detail caption construction.
GLaMM~\cite{rasheed2023glamm} and ASMv2~\cite{wang2024all} use open-source suites for dense caption generation, with a focus on correspondence of local descriptions and image regions. 
Our proposed data construction pipeline adopts a divide-and-conquer strategy, unleashing LVLM's detail caption ability by generating and merging local captions. 
A recent work Monkey~\cite{li2023monkey} also adopts a zoom-in-and-caption approach, but they use outdated local captioner and rely on ChatGPT for caption generation. 
Compared with Monkey, we use open-source LVLM and LLM to synthesize detail caption data, and propose a phrase-level filtering strategy. 
Guided by the proposed benchmark, we also provide in-depth analysis for the effectiveness of the detail caption construction pipeline.

\section{Benchmarking Detail Image Caption}
\label{sec: CAPTURE}
In this section, we elaborate the expert judgement data construction process and the workflow of the proposed detail image caption metric. 

\subsection{Detail Caption Evaluation Datasets}
\label{sec: 3.1}
\begin{table}[h]
\caption{
Statistics of our DetailCaps-100 and DetailCaps-4870 benchmark. 
``Annt. expert” means the source
``Ref num” indicates the number of reference captions. 
``Uni. 2-gram” denotes the unique 2-gram number in reference captions.  
}
\vspace{-1mm}
\resizebox{\textwidth}{12.5mm}{
\begin{tabular}{l | l c c c c c}
\toprule
Benchmark & Data source & Annt. expert & Img num & Ref num & Avg len & Uni. 2-gram
\\
\midrule
$\text{COCO}_{test}$       & COCO~\cite{lin2014microsoft} & Human & $5000$ & $25,010$ & $10.59$ & $61,448$ \\
$\text{Nocaps}_{val}$       & Openimages~\cite{krasin2017openimages} & Human & $4500$ & $45,000$ & $11.49$ & $116,969$ \\
\midrule
\multirow{2}{*}{DetailCaps-100}   & COCO~\cite{lin2014microsoft}, SAM~\cite{kirillov2023segment} & \multirow{2}{*}{Human} & \multirow{2}{*}{$100$}  & \multirow{2}{*}{$100$}& \multirow{2}{*}{$175.96$} & \multirow{2}{*}{$10,858$} \\
& LAION~\cite{schuhmann2021laion}, CC~\cite{sharma2018conceptual}, SBU~\cite{ordonez2011im2text} & & & & \\
\multirow{2}{*}{DetailCaps-4870}  & COCO~\cite{lin2014microsoft}, SAM~\cite{kirillov2023segment}, LAION~\cite{schuhmann2021laion} & GPT-4V, GPT-4O  & \multirow{2}{*}{$4870$} & \multirow{2}{*}{$14610$} & \multirow{2}{*}{$122.06$} & \multirow{2}{*}{$533,201$} \\
& CC~\cite{sharma2018conceptual}, SBU~\cite{ordonez2011im2text}, Coyo~\cite{kakaobrain2022coyo-700m}, Flickr~\cite{young-etal-2014-image} & Gemini-1.5-Pro &  & & & \\
\bottomrule
\end{tabular}
}
\centering
\vspace{-1mm}
\label{tbl: bench_statistics}
\end{table}
To benchmark detail image caption task reliably and better evaluate the consistency between each image caption metric and expert evaluation, we construct two expert-annotated datasets for performance evaluation.

For human evaluation dataset, we curate 100 cases sampled from ShareGPT4V-102k~\cite{chen2023sharegpt4v} randomly. 
We first call GPT-4V to generate detail captions, followed by human experts removing hallucinations and supplementing omitted visual elements. 
The refined detail image captions are then used as the ground truth for evaluation. 
We prompt three LVLMs with leading detail captioning performance for caption generation, which are ShareCaptioner~\cite{chen2023sharegpt4v}, CogVLM~\cite{Wang2023CogVLMVE} and LLaVA-1.5~\cite{liu2023improved}.  
Human experts are instructed to score each caption based on the precision and recall of three types of visual elements: object, attribute and relation. 
The overall scores range in $[0, 5]$, and are normalized to $[0, 1]$ for fair expert judgement consistency evaluation of caption metrics. 

We further curate a 4870 case dataset annotated by GPT-4V, Gemini-1.5-Pro and GPT-4O for detail caption evaluation. 
Besides the data sources used in human-annotated 100 cases, we further incorporate pictures from COYO~\cite{kakaobrain2022coyo-700m}, LAION~\cite{schuhmann2021laion}, CC~\cite{changpinyo2021conceptual} and Flickr~\cite{young2014image} for diversity. 
Captions generated by ShareCaptioner, CogVLM and LLaVA-1.5 and corresponding annotated caption scores are provided for each sample. 
We instruct text-only GPT-4\cite{achiam2023gpt} to compare model-generated captions with GPT-4V, Gemini-1.5-Pro and GPT-4O annotated references to obtain evaluation scores.
We use text-only GPT-4 for evaluation because of its outstanding instruction following abilities. 
We refer to Appendix~\ref{sec: app.prompt} for more details about the prompts used for detail caption generation and GPT4 evaluation generation.

We show the statistics of the curated expert judgement datasets in Table~\ref{tbl: bench_statistics}. 
Our detail caption evaluation benchmarks contain image samples from various sources, and the reference captions are significantly longer than previous benchmarks. 
It worth noticing that DetailCaps-4870 benchmark contains 377,184 unique 2-grams in 9740 reference captions, while has only 116,969 unique 2-grams across 45,000 references.

\subsection{CAPTURE Metric}
CAPTURE metric extracts and matches core visual elements instead of \textit{n}-gram pieces to obtain evaluation results, suppressing the influence of varying writing styles. 
We elaborate the design of CAPTURE metric in the following parts: visual elements extraction, stop words filtering and visual elements matching. 
We refer to Appendix~\ref{sec: app.capture} for implementation details of CAPTURE metric.

\begin{figure*}[t]
\centering
\begin{tikzpicture}
\draw (0,0 ) node[inner sep=0] {\includegraphics[width=\columnwidth, trim={0.5cm 0cm 00.2cm 0cm}, clip]{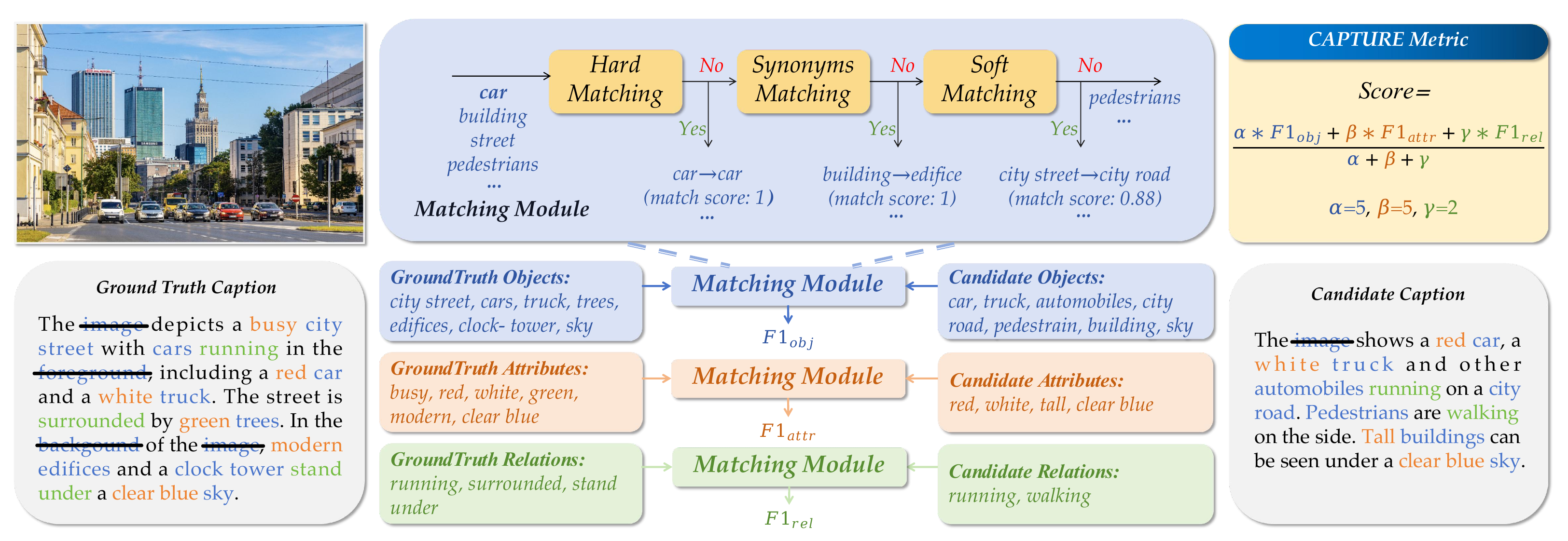}};
\end{tikzpicture}
\caption{
An illustration of the proposed detail caption evaluation metric CAPTURE.
The crossed text indicates objects discarded by the stop words filtering module. 
}\label{fig: detaicaps_method}
\vspace{-3mm}
\end{figure*}

\paragraph{Visual elements extraction. }
Visual elements extraction module extracts objects, attributes and relations from caption sentences. 
We adopt Factual parser~\cite{li-etal-2023-factual}, which is a T5-base model with leading performance in text scene graph parsing. 
Since Factual parser is trained on short caption parsing dataset, we use NLTK toolkit~\cite{bird2006nltk} to split detail image caption into sentences to be parsed separately. 
The parsing results are then lemmatized (Wordnet~\cite{miller1995wordnet}), deduplicated and merged to be the final parsing result.

\paragraph{Stop words filtering. }
Factual parser may extract abstract nouns as object elements, for example "foreground", "background", which do not correspond to visual elements in the image, and are not expected to participate in the matching process.
To this end, we curate a stop word list to filter out these abstract nouns from extracted object elements. 
We first apply LLaMA2-13B-chat~\cite{touvron2023llama} and Factual parser to ShareGPT4V-102k dataset for nouns extraction respectively, and curate words recalled by Factual parser but omitted by LLaMA2-13B-chat. 
We compute the frequency of these words and task human experts to judge whether words with the highest frequencies have tangible meanings. 
Finally, 317 words with high frequency are included in the stop word list.

\paragraph{Visual elements matching. }
In this part, we match the extracted visual elements to produce evaluation result. 
We implement a three-stage matching strategy to obtain matching results, which is robust to varying writing styles. 
An illustration of the matching module is shown in Figure~\ref{fig: detaicaps_method}. 
We first match the same visual elements, followed by a synonym matching module. 
Words sharing one or more synonyms are considered matched, where 
Wordnet~\cite{miller1995wordnet} is employed to get the synonym set of visual elements. 
Phrases matched in exact or synonym matching module obtain a $1.0$ matching score. 
To deal with the remaining unmatched elements,
we further propose a soft matching module, which uses Sentence BERT~\cite{devlin2018bert} model to compute soft matching score.
To be specific, we use Sentence BERT to encode the remaining object, attribute and relation phrases and compute the cosine similarity matrix between ground truth phrase embeddings and candidate ones. 
The max similarity score of each row and column, which is in $[0.0, 1.0)$, are the added up to the exact matching and synonym matching scores. 
We then compute the precision, recall and F1 of visual elements based on the matching score. 
CAPTURE metric computes the caption quality score as a weighted summation of the three F1 scores, which is illustrated in Figure~\ref{fig: detaicaps_method}.
We set weights for each type of visual elements as Object:Attribute:Relation=5:5:2 by default.

\section{Improving Detail Image Caption}
\label{sec: detailcap}
In this section, we elaborate the design of the proposed detail caption synthesizing pipeline, and introduce how to improve LVLM training with constructed detail caption data.

\subsection{Detail Caption Construction}
\label{sec: detail_cap_con}
We introduce the proposed divide-and-conquer detail caption construction pipeline in the following five stages. 
The pipeline is illustrated in the right part of Figure~\ref{fig: intro}. 


\paragraph{Stage I: Overall caption generation. }
We first instruct a given LVLM to generate overall image caption as the skeleton for high quality detail caption generation. 
The overall caption may suffer from hallucinations and omissions, and will be polished in the following stages.

\paragraph{Stage II: Salient visual elements detection. }
To locate salient objects for local caption generation, we segment the image with SAM~\cite{kirillov2023segment} and filter out masks with extreme large or small sizes. 
Then, we adopt a maximal rectangle algorithm to reduce overlap between remaining masks.
The resulted cropped bounding boxes are regarded as salient visual elements. 

\paragraph{Stage III: Local caption generation. }
To produce complementary detail visual information for the overall caption, we instruct the given LVLM to generate local caption for each bounding box obtained in Stage II.
We limit the output length of local captions to be no more than twenty words to suppress unexpected hallucinations.

\paragraph{Stage IV: Hallucination filtering. }
We propose a novel phrase-level filtering strategy to suppress hallucinations and preserve the recalled visual elements.
We first extract visual element phrases from both overall caption and local captions with Factual parser, and filter out those scored lower than 0.01 by Owlv2~\cite{minderer2024scaling}, which is an open-vocabulary object detection model. 
Notice that captions may suffer from some grammar errors with phrases filtered out.
These errors will be corrected in the final stage. 

\paragraph{Stage V: Caption merging. }
In this stage, an LLM is instructed to merge local captions into the skeleton provided in the overall caption smoothly, instead of simply concatenating them. 

With local caption providing supplementary visual information and filtering module tackling accompanied hallucinations, the synthesized detail image caption captures more visual elements with hallucinations suppressed. 
Visualized examples are shown in Appendix \ref{sec: app.visualization}.

\subsection{Improving LVLM Training with Synthesized Detail Caption Data}
We further explore to enhance LVLM's overall understanding performance with self-generated detail caption data. 
We synthesize detail caption data for images from ShareGPT4V-102k dataset~\cite{chen2023sharegpt4v}, and then select a proportion of synthesized detail caption data for model training. 
Samples with the largest number of visual elements extracted by Factual parser are selected for their rich visual information.
The selected data is incorporated into the SFT dataset to improve overall understanding performance.

\section{Experiments}
\label{sec: exp}
In this section, we introduce the experiment settings and show main experimental results to demonstrate the effectiveness of the proposed detail image caption metric and data construction pipeline. 

\subsection{Benchmarking Detail Image Caption}

\subsubsection{Experiment Settings}

\paragraph{Datasets. }
We conduct experiments on the two expert judgement datasets described in Section~\ref{sec: 3.1}.
Each sample in the two datasets contains expert-annotated reference detail captions, and expert-annotated caption quality scores for three SOTA LVLM-generated captions. 
The statistics of the two datasets are shown in Table~\ref{tbl: bench_statistics}.

\paragraph{Evaluation protocol. }
We evaluate the caption metrics' consistency with expert judgements with four metrics: Pearson correlation coefficient (PCC) $\rho$, coefficient of determination $R^2$, Kendall's $\tau$ (Kd $\tau$) and Sample $\tau$ (Sp $\tau$).
PCC reflects the linear correlation between the metric-evaluated scores and the expert-annotated ones. 
Coefficient of determination evaluates both the linear correlation and the variation of metric-evaluated score values from expert judgement.
Kd $\tau$ is computed as the proportion of matched score order pairs among all partial order pairs. 
Sp $\tau$ computes Kd $\tau$ for each sample's caption scores independently, and use the average value as final result.
Sp $\tau$'s formulation fits LVLM's caption evaluation process well, and therefore is regarded as the most important metric for consistency evaluation. 

\paragraph{Baselines. }
We compare the CAPTURE metric with both rule-based and model-based caption metrics.
BLEU-2~\citep{papineni2002bleu}, CIDER~\cite{vedantam2015cider}, ROUGE-L~\cite{lin2004rouge} and METEOR~\cite{banerjee2005meteor} are considered as representative rule-based metrics. 
For model-based metrics, we consider SPICE~\cite{anderson2016spice}, CLIPScore~\cite{hessel-etal-2021-clipscore} and PAC-S~\cite{sarto2023positive}.
SPICE is built on a PCFG text parser model for information extraction, while CLIPScore and PAC-S borrow CLIP model to evaluate the alignment between images and text captions. 
We implement the model-based metrics with OpenCLIP-L/14~\cite{cherti2023reproducible}, and truncate the detail caption paragraph for alignment score computation due to the limitation in input length. 
We also evaluate the consistency between GPT-Eval and human judgements on DetailCaps-100 benchmark. 

\subsubsection{Main Results}

\begin{table}[t]
\vspace{-3mm}
\caption{
Image caption metrics' evaluation consistency with expert judgements. 
Bold number indicates the best result among all caption metrics. 
Italic numbers indicate GPT-EVAL results. 
}
\resizebox{\textwidth}{20.5mm}{
\begin{tabular}{l cccc cccc cccc}
\toprule
\multicolumn{1}{c}{Metric} &
\multicolumn{4}{c}{DetailCaps-100} &
\multicolumn{4}{c}{DetailCaps-4870} &
\multicolumn{4}{c}{Average} 
\\
\cmidrule(lr{0.5em}){2-5}\cmidrule(lr{0.5em}){6-9}\cmidrule(lr{0.5em}){10-13}
& PCC $\rho$ & $1-R^2$ & Kd $\tau$ & Sp $\tau$ & PCC $\rho$ & $1-R^2$ & Kd $\tau$ & Sp $\tau$ & PCC $\rho$ $\uparrow$ & $1-R^2$ $\downarrow$ & Kd $\tau$ $\uparrow$ & Sp $\tau$ $\uparrow$ \\
\midrule
\rowcolor{gray!20}
\multicolumn{13}{c}{\textit{Rule-based metrics}} \\
\textsc{BLEU}~\citeyearpar{papineni2002bleu}       & $0.2150$ & $96.27$ & $0.1623$ & $0.2163$ & $0.3066$ & $13.24$ & $0.2109$ & $0.2760$ & $0.2608$ & $54.75$ & $0.1866$ & $0.2462$ \\
\textsc{ROUGE}~\citeyearpar{lin2004rouge}        & $0.2554$ & $185.69$ & $0.1905$ & $0.3321$ & $0.3347$ & $82.55$ & $0.2393$ & $0.3445$ & $0.2951$ & $134.12$ & $0.2149$ & $0.3383$ \\
\textsc{METEOR}~\citeyearpar{banerjee2005meteor}        & $0.3643$ & $384.58$ & $0.2679$ & $0.3529$ & $0.4400$ & $196.19$ & $0.3175$ & $0.4594$ & $0.4022$ & $290.38$ & $0.2927$ & $0.4062$ \\
\textsc{CIDER}~\citeyearpar{vedantam2015cider}        & $0.0834$ & $1.7e^{7}$ & $0.1159$ & $0.0564$ & $0.1462$ & $3.5e^{7}$ & $0.1171$ & $0.1418$ & $0.1148$ & $2.6e^{7}$ & $0.1165$ & $0.0991$ \\
\rowcolor{gray!20}
\multicolumn{13}{c}{\textit{Model-based metrics}} \\
\textsc{SPICE}~\citeyearpar{anderson2016spice}     & $0.3580$ & $126.60$ & $0.2641$ & $0.3819$ & $0.5192$ & $185.30$ & $0.3847$ & $0.5616$ & $0.4386$ & $155.95$ & $0.3244$ & $0.4718$ \\
\textsc{RefCLIPScore}~\citeyearpar{hessel-etal-2021-clipscore}     & $0.2538$ & $31.82$ & $0.1829$ & $0.3244$ & $0.4577$ & $11.11$ & $0.3129$ & $0.4437$ & $0.3558$ & $21.46$ & $0.2479$ & $0.3841$ \\
\textsc{RefPAC-S}~\citeyearpar{sarto2023positive}       & $0.2664$ & $60.67$ & $0.1946$ & $0.3221$ & $0.4135$ & $19.95$ & $0.3246$ & $0.3825$ & $0.3399$ & $40.31$ & $0.2596$ & $0.3523$ \\
\midrule
\textsc{CAPTURE}       & $\bm{0.4735}$ & $\bm{11.58}$ & $\bm{0.3688}$ & $\bm{0.6117}$ & $\bm{0.5446}$ & $\bm{5.00}$ & $\bm{0.4033}$ & $\bm{0.5919}$ & $\bm{0.5091}$ & $\bm{8.29}$ & $\bm{0.3861}$ & $\bm{0.6018}$ \\
\textsc{GPT4-Eval}       & $\emph{0.5157}$ & $\emph{44.44}$ & $\emph{0.4237}$ & $\emph{0.6120}$ & $-$ & $-$ & $-$ & $-$ & $-$ & $-$ & $-$ & $-$ \\

\bottomrule
\end{tabular}
}
\centering
\vspace{0mm}
\label{tbl: capture_main}
\vspace*{-1mm}
\end{table}

\paragraph{CAPTURE achieves the highest consistency with expert judgements. }
As shown in Table~\ref{tbl: capture_main}, the proposed metric CAPTURE improves PCC $\rho$ by $0.0683$ ($15.6\%\uparrow$), $R^2$ score by $24.26$ ($74.7\%\downarrow$), Kd $\tau$ by $0.0592$ ($18.3\%\uparrow$) and Sp $\tau$ by $0.1240$ ($26.4\%\uparrow$) over previous SOTA baselines. 
The advantages in PCC $\rho$, Kd $\tau$ and Sp $\tau$ indicate that the proposed metric performs the best in linear correlation with expert judgment and pair-wise ranking accuracy, showing promising prospects for LVLM-generated detail caption evaluation.
Besides, CAPTURE also performs the best in $1-R^2$ metric, indicating that CAPTURE produces evaluation results with aligned values.  

\paragraph{METEOR and SPICE perform the best among rule-based and model-based metrics, respectively.}
We attribute METEOR's satisfying performance to its consideration for both precision and recall of \textit{n}-grams.
METEOR also adopts exact, synonym and porter stem matching strategies, improving its robustness to varying writing styles.
For SPICE, its PCFG parser performs more robust for long detail captions compared with CLIP-based metrics, which suffer from CLIP's limited input text length. 

\paragraph{GPT4-Eval achieves the highest consistency with human evaluation on DetailCaps-100 dataset.}
This result validates the effectiveness of evaluating caption metrics' consistency with GPT4-Eval results on the larger dataset DetailCaps-4870. 
It is also worth noticing that CATURE's consistency performance is pretty close to that of GPT-Eval.
Moreover, CAPTURE does not require calling expensive LLM APIs, demonstrating its promising prospect in detail caption evaluation.

\subsubsection{Analysis}
We verify the effectiveness of the design of CAPTURE metric. 
Among the consistency evaluation metrics, we point out that Sp $\tau$ is the closest to real detail caption evaluation scenario, and we focus on this metrics for analysis.

\paragraph{Stop words filtering improves sample-level evaluation consistency effectively.}
Statistics show that when evaluating candidate captions on DetailCaps-100 dataset, 28.43\% extracted object phrases are detected and discarded by the stop words filtering module.
As shown in Table~\ref{tbl: capture_ablation}, performance drops on SP $\tau$ are witnessed on both DetailCaps-100 and DetailCaps-4870 benchmark when stop words filtering module is removed. 
We attribute the fluctuation in other consistency metrics to the varying number of visual elements discarded by the stop words filtering module across samples. 


\paragraph{Soft matching module improves evaluation consistency and the alignment of evaluation score values.}
When soft matching module is removed, CAPTURE suffers from a $3.3\%\downarrow$ performance drop in Sp $\tau$.
It is also worth noticing that $1-R^2$ score deteriorates the most significantly. 
The soft matching strategy tackles a variety of phrases with similar meaning, and thus makes up the deficiency of exact matching and synonym matching modules when tackling varying writing styles. 

\paragraph{The default $\alpha,\beta,\gamma=5,5,2$ setting is a sweet spot for detail caption evaluation.}
We modify the scale factors of relation elements $\gamma$ from $0$ (discarding relation matching score) to $5$ (relation F1 is considered equally with object F1 and attribute F1) to verify this judgement. 
Experiment results show that CAPTURE's performance drops with relation matching score ratio $\gamma$ as $0$ or $5$, validating that $\alpha,\beta,\gamma=5,5,2$ is the most suitable for CAPTURE's evaluation. 

\begin{table}[t]
\caption{
Ablation study for the design of CAPTURE score. 
We demonstrate the effectiveness of the proposed stop words filtering and soft matching module, and validate that CAPTURE's default setting $\alpha,\beta,\gamma=5,5,2$ is a sweet spot for detail caption evaluation.
}
\resizebox{\textwidth}{15mm}{
\begin{tabular}{l cccc cccc cccc}
\toprule
\multicolumn{1}{c}{Metric} &
\multicolumn{4}{c}{DetailCaps-100} &
\multicolumn{4}{c}{DetailCaps-4870} &
\multicolumn{4}{c}{Average} 
\\
\cmidrule(lr{0.5em}){2-5}\cmidrule(lr{0.5em}){6-9}\cmidrule(lr{0.5em}){10-13}
& PCC $\rho$ & $1-R^2$ & Kd $\tau$ & Sp $\tau$ & PCC $\rho$ & $1-R^2$ & Kd $\tau$ & Sp $\tau$ & PCC $\rho$ $\uparrow$ & $1-R^2$ $\downarrow$ & Kd $\tau$ $\uparrow$ & Sp $\tau$ $\uparrow$ \\
\midrule
\textsc{CAPTURE}       & $0.4735$ & $11.58$ & $0.3688$ & $0.6117$ & $0.5446$ & $5.00$ & $0.4033$ & $0.5919$ & $0.5091$ & $8.29$ & $0.3861$ & $0.6018$ \\
\quad \textsc{- Stop words}       & $0.4830$ & $13.23$ & $0.3804$ & $0.5947$ & $0.5456$ & $6.13$ & $0.4047$ & $0.5859$ & $0.5143$ & $9.68$ & $0.3926$ & $0.5903$ \\
\quad \textsc{- Soft matching}       & $0.4674$ & $29.15$ & $0.3488$ & $0.5770$ & $0.5616$ & $20.35$ & $0.4116$ & $0.5914$ & $0.5145$ & $24.75$ & $0.3802$ & $0.5842$ \\
\quad \textsc{$\alpha,\beta,\gamma=5,5,0$}       & $0.4654$ & $9.21$ & $0.3642$ & $0.5947$ & $0.5335$ & $4.05$ & $0.4002$ & $0.5802$ & $0.4994$ & $6.63$ & $0.3822$ & $0.5875$ \\
\quad \textsc{$\alpha,\beta,\gamma=5,5,5$}       & $0.4651$ & $13.75$ & $0.3556$ & $0.6064$ & $0.5388$ & $5.92$ & $0.3936$ & $0.5844$ & $0.5020$ & $9.84$ & $0.3746$ & $0.5954$ \\
\quad \textsc{$\alpha,\beta,\gamma=3,7,2$}       & $0.4842$ & $10.59$ & $0.3863$ & $0.5654$ & $0.5308$ & $5.17$ & $0.3993$ & $0.5846$ & $0.5075$ & $7.88$ & $0.3928$ & $0.5750$ \\
\quad \textsc{$\alpha,\beta,\gamma=7,3,2$}       & $0.4384$ & $10.10$ & $0.3458$ & $0.6010$ & $0.5231$ & $4.36$ & $0.3874$ & $0.5698$ & $0.4808$ & $7.23$ & $0.3666$ & $0.5854$ \\

\bottomrule
\end{tabular}
}
\centering

\label{tbl: capture_ablation}
\end{table}

\begin{table}[t]
\caption{
CAPTURE scores of open source models on DetailCaps-100 ($\text{DC}_{100}$) and DetailCaps-4870 ($\text{DC}_{4870}$) benchmarks. 
``Annt.” indicates how the detail caption data is annotated. 
}
\resizebox{1.0\textwidth}{24mm}{
\begin{tabular}{l l l l | c c}
\toprule
LVLM &  Language & Detail Caption Data & Resolution & $\text{DC}_{100}$ & $\text{DC}_{4870}$
\\

\midrule
\textsc{CogVLM\citeyearpar{Wang2023CogVLMVE}}       &  Vicuna-7B & Human Annt.  & $490^2$ & $63.01$ & $60.06$ \\
\textsc{ShareCaptioner-7B\citeyearpar{chen2023sharegpt4v}}       &  Vicuna-7B & GPT-4V Annt. & $448^2$ & $60.85$ & $59.80$ \\
\textsc{LLaVA-1.5-7B\citeyearpar{liu2023improved}}       &  Vicuna-7B & Synthesized & $336^2$ & $51.23$ & $51.05$ \\
\textsc{LLaVA-1.5-13B\citeyearpar{liu2023improved}}       &  Vicuna-13B & Synthesized & $336^2$ & $51.74$ & $51.20$ \\
\textsc{LLaVA-NEXT-7B\citeyearpar{liu2024llavanext}}      &  Vicuna-7B & GPT-4V Annt. & $336^2$*\{$1$\text{-}$5$\} & $60.18$ & $58.61$ \\
\textsc{LLaVA-NEXT-13B\citeyearpar{liu2024llavanext}}      &  Vicuna-13B & GPT-4V Annt. & $336^2$*\{$1$\text{-}$5$\} & $60.38$ & $59.01$ \\
\textsc{LLaVA-NEXT-34B\citeyearpar{liu2024llavanext}}      & Hermes-2-Yi-34B & GPT-4V Annt. & $336^2$*\{$1$\text{-}$5$\} & $60.60$ & $59.20$ \\
\textsc{Mini-Gemini-HD-7B\citeyearpar{li2024mini}}       &  Vicuna-7B & GPT-4V Annt. & $336^2$*$5$ & $59.51$ & $57.95$ \\
\textsc{Mini-Gemini-HD-13B\citeyearpar{li2024mini}}       &  Vicuna-13B & GPT-4V Annt. & $336^2$*$5$ & $60.51$ & $58.66$ \\
\textsc{Intern-XComposerV2\citeyearpar{dong2024internlm}}       &  Vicuna-7B & GPT-4V Annt. & $490^2$ & $61.43$ & $59.86$ \\
\textsc{InternVL-V1.2-PLUS-40B\citeyearpar{chen2023internvl}}       &  Hermes-2-Yi-34B & GPT-4V Annt. & $448^2$ & $61.61$ & $60.69$ \\
\textsc{InternVL-V1.5-26B\citeyearpar{chen2024far}}       &  InternLM-20B & GPT-4V Annt. & $448^2$*\{$1$\text{-}$41$\} & \bm{$65.62$} & \bm{$63.42$} \\
\bottomrule
\end{tabular}
}
\centering
\label{tbl: capture_scores}
\vspace*{-3mm}
\end{table}

\subsubsection{Evaluating LVLMs with Leading Performance}
With DetailCaps benchmark and CAPTURE evaluating LVLMs' detail captioning performance reliably, we review the detail caption capabilities for 12 open source LVLMs with leading performance.
The evaluation results on DetailCaps-100 and DetailCaps-4870 are shown in Table~\ref{tbl: capture_scores}. 
Among all models, InternVL-V1.5~\cite{chen2024far} achieves the best detail image caption performance with a large advantage over other models. 
It also can be observed from the results of the LLaVA-1.5, LLaVA-Next and Mini-Gemini\cite{li2024mini} series that model's detail captioning ability improves consistently as the model size increases. 
In addition, a common observation is that training with detail caption data generated by GPT-4V leads to better detail captioning performance. 
Among these LVLMs, CogVLM achieves the second highest CAPTURE score with high-quality human-refined detail image caption data.

\subsection{Improving Detail Image Caption}

\subsubsection{Experiment Settings}
We use ShareGPT4V-102k dataset for detail caption data construction and implement two pipelines with different model size. 
For 7B model pipeline, we use SAM-ViT-L~\cite{kirillov2023segment} for segmentation, LLaVA-1.5-7B for overall and local caption generation, OwlV2-large-ensemble~\cite{minderer2024scaling} for hallucination filtering and LLaMA-2-7B-Chat for caption mering. 
For 13B model pipeline, we replace SAM-ViT-H, LLaVA-1.5-13B, and LLaMA-2-13B-Chat instead.
We validate the effectiveness of the proposed data construction pipeline with four LVLMs with leading performance, which are LLaVA-1.5-7B, LLaVA-1.5-13B, LLaVA-NEXT-7B and Mini-Gemini-7B-HD. 

\begin{table}[ht]
\caption{
Performance improvement of the proposed detail caption synthesizing pipeline with SOTA LVLMs as backbones.
Overall precision and recall are computed as a weighted sum of each type of visual element's score. 
The weights are set according to CAPTURE's $\alpha,\beta,\gamma=5,5,2$ setting. 
``Self” indicates detail caption generated by LVLM directly, and ``Synthesized” means data is constructed through our five-stage pipeline.
}
\vspace{-1mm}
\resizebox{\textwidth}{24mm}{
\begin{tabular}{l ccc ccc ccc}
\toprule
\multicolumn{1}{c}{Caption} &
\multicolumn{3}{c}{Detailcaps-100} & 
\multicolumn{3}{c}{Detailcaps-4870} & 
\multicolumn{3}{c}{Average} 
\\
\cmidrule(lr{0.5em}){2-4}\cmidrule(lr{0.5em}){5-7}\cmidrule(lr{0.5em}){8-10}
& CAPTURE & Precision & Recall & CAPTURE & Precision & Recall & CAPTURE & Precision & Recall \\
\midrule
\rowcolor{gray!20}
\multicolumn{10}{c}{\textit{LLaVA-1.5-7B}} \\
\textsc{Self}       & $51.23$ & $65.24$ & $43.31$ & $51.05$ & $65.77$ & $43.04$ & $51.14$ & $65.50$ & $43.17$ \\ 
\textsc{Synthesized}       & $57.11$ & $64.12$ & $52.08$ & $56.25$ & $64.35$ & $50.79$ & $56.68$ & $64.23$ & $51.44$ \\
\rowcolor{gray!20}
\multicolumn{10}{c}{\textit{LLaVA-1.5-13B}} \\
\textsc{Self}       & $51.76$ & $65.01$ & $44.10$ & $51.20$ & $66.25$ & $43.13$ & $51.48$ & $65.63$ & $43.62$ \\
\textsc{Synthesized}       & $57.36$ & $62.07$ & $53.52$ & $57.05$ & $62.98$ & $52.67$ & $57.20$ & $62.52$ & $53.09$ \\

\rowcolor{gray!20}
\multicolumn{10}{c}{\textit{LLaVA-NEXT-7B}} \\
\textsc{Self}       & $61.48$ & $65.60$ & $57.82$ & $58.61$ & $65.60$ & $55.75$ & $60.73$ & $65.60$ & $56.78$ \\
\textsc{Synthesized}       & $62.24$ & $64.49$ & $60.07$ & $60.39$ & $63.82$ & $57.85$ & $61.31$ & $64.16$ & $58.96$ \\

\rowcolor{gray!20}
\multicolumn{10}{c}{\textit{Mini-Gemini-7B-HD}} \\
\textsc{Self}       & $59.51$ & $61.99$ & $57.28$ & $57.95$ & $61.56$ & $55.25$ & $58.73$ & $61.78$ & $56.27$ \\
\textsc{Synthesized}       & $60.44$ & $60.98$ & $59.78$ & $59.07$ & $60.16$ & $58.60$ & $59.75$ & $60.57$ & $59.19$ \\

\bottomrule
\end{tabular}
}
\centering
\label{tbl: detailcaps_main}
\vspace{-1mm}
\end{table}

\subsubsection{Main results}

\paragraph{Our detail caption synthesizing pipeline improves LVLM-generated caption quality effectively.}
As shown in Table~\ref{tbl: detailcaps_main}, for LLaVA-1.5-7B and LLaVA-1.5-13B, the detail caption quality is improved by a large fraction in terms of CAPTURE score. 
For more advanced LVLM like LLaVA-NEXT and Mini-Gemini-HD, the advantage of the proposed pipeline persists, demonstrating the effectiveness of the our data synthesizing strategy. 
We attribute the smaller fraction of improvement in LLaVA-NEXT and Mini-Gemini-HD to other vision and language tools' limited capabilities, which pose "short boards" compared with LVLMs trained with expert-annotated detail caption training data.

\paragraph{Our pipeline enhances recall of visual elements effectively with little precision drop.}
As shown in Table~\ref{tbl: detailcaps_main}, this tendency can be observed across all four LVLMs, indicating that the divide-and-conquer strategy improves model's perception of detail visual elements effectively. 
Thanks to the hallucination filtering module, the performance drop in precision is suppressed, so that improvement on CAPTURE score is witnessed across all LVLMs.

\subsubsection{Analysis}

\begin{table}[t]
\caption{
Analysis for LLaVA-1.5-7B's detail image caption performance in terms of CAPTURE score. 
We investigate the influence of different hallucination filtering methods, and demonstrate the effectiveness of the proposed self-looping strategy. 
We refer to Table~\ref{tbl: detailcaps_main} for definitions of terms.
}
\resizebox{\textwidth}{23.5mm}{
\begin{tabular}{l ccc ccc ccc}
\toprule
\multicolumn{1}{c}{Caption} &
\multicolumn{3}{c}{Detailcaps-100} & 
\multicolumn{3}{c}{Detailcaps-4870} & 
\multicolumn{3}{c}{Average} 
\\
\cmidrule(lr{0.5em}){2-4}\cmidrule(lr{0.5em}){5-7}\cmidrule(lr{0.5em}){8-10}
& CAPTURE & Precision & Recall & CAPTURE & Precision & Recall & CAPTURE & Precision & Recall \\
\midrule
\rowcolor{gray!20}
\multicolumn{10}{c}{\textit{Ablation}} \\
\textsc{Self}       & $51.23$ & $65.24$ & $43.31$ & $51.05$ & $65.77$ & $43.04$ & $51.14$ & $65.50$ & $43.17$ \\ 
\textsc{Synthesized}       & $57.11$ & $66.31$ & $52.16$ & $56.25$ & $64.35$ & $50.79$ & $56.68$ & $64.23$ & $51.44$ \\
\quad \textsc{- filter}       & $56.78$ & $65.16$ & $53.26$ & $56.08$ & $64.09$ & $50.81$ & $56.43$ & $64.62$ & $52.03$ \\
\quad \textsc{vqa filter}       & $56.44$ & $63.95$ & $51.11$ & $55.89$ & $64.07$ & $50.41$ & $56.16$ & $64.01$ & $50.76$ \\
\quad \textsc{filter local}       & $56.75$ & $63.87$ & $51.61$ & $56.34$ & $64.06$ & $51.22$ & $56.55$ & $63.97$ & $51.41$ \\

\rowcolor{gray!20}
\multicolumn{10}{c}{\textit{Self-looping}} \\
\textsc{Self}       & $51.23$ & $65.24$ & $43.31$ & $51.05$ & $65.77$ & $43.04$ & $51.14$ & $65.50$ & $43.17$ \\ 
\quad \textsc{loop1}       & $51.91$ & $63.48$ & $45.02$ & $52.35$ & $64.98$ & $45.03$ & $52.13$ & $64.23$ & $45.03$ \\
\quad \textsc{loop2}       & $52.50$ & $63.43$ & $45.66$ & $52.43$ & $63.81$ & $45.70$ & $52.47$ & $63.62$ & $45.68$ \\
\quad \textsc{loop3}       & $52.89$ & $62.45$ & $46.86$ & $52.78$ & $63.00$ & $46.52$ & $52.84$ & $62.73$ & $46.69$ \\
\quad \textsc{loop4}       & $54.02$ & $62.24$ & $48.45$ & $54.37$ & $62.89$ & $48.78$ & $54.20$ & $62.56$ & $48.62$ \\

\bottomrule
\end{tabular}
}
\centering
\vspace{0.5mm}
\label{tbl: detailcaps_analysis}
\vspace{-3mm}
\end{table}

\paragraph{Our phrase-level hallucination filtering strategy achieves the best performance.}
As shown in Table~\ref{tbl: detailcaps_analysis}, when the filtering module is removed (-filter), a performance drop in CAPTURE score is witnessed. 
We also compare our filtering strategy with other alternatives used in Monkey~\cite{li2023monkey}.
For VQA filtering, we use LVLM to check if the visual element phrase exists in the image. 
For local caption filtering, we filter out hallucinated local caption sentences rather than extracted phrases.
Experiment results show that both alternatives lead to performance drops in CAPTURE score, demonstrating the effectiveness of the proposed phrase-level filtering strategy. 

\paragraph{LVLM's detail caption ablity can be improved via self-looping.}
We adopt LLaVA-1.5-7B as the backbone LVLM, and synthesize detail caption data for model's training. 
In each loop, we rerun the SFT stage of LLaVA-1.5-7B from a pretrain checkpoint (without any SFT), with annotated 25k detail caption data incorporated into the training data. 
Experiment results are shown in Table~\ref{tbl: detailcaps_analysis}.
Model's detail captioning ability keeps improving in the listed 4 loops, showing a promising self-evolutioning phenomena in detail captioning performance. 


\subsubsection{Improving LVLM Training with Synthesized Detail Caption Data}
\paragraph{Experiment Settings. }
We follow LLaVA-1.5~\cite{liu2023improved} pipeline for model training. 
The vision-language projector is trained with 558k short caption data and a 128 batch size during pretraining, and all parameters except the vision module are trained with 665k visual instruction tuning data and a 256 batch size during SFT.
We train the model with AdamW optmizer, with a $1e^{-4}$ pretraining learning rate and a $2e^{-5}$ SFT learning rate. 
We add 25k detail caption data into SFT stage for the 7B model, and 50k for the 13B model due to its larger capacity. 
In our experiments, the pretraining process takes 24 GPU hours and SFT takes 88 GPU hours on Nvidia A100. 
We use MME~\cite{fu2023mme}, MMMU~\cite{yue2023mmmu}, MMStar~\cite{chen2024we}, GQA~\cite{hudson2019gqa}, VizWiz~\cite{gurari2018vizwiz}, POPE~\cite{li2023evaluating} and the proposed DetailCaps benchmarks for model's natural scene visual understanding ability evaluation. 
RefCOCOg~\cite{mao2016generation} is a referring expression comprehension task to evaluate model's detail understanding capability. 
OCRBench~\cite{liu2023hidden} and DocVQA~\cite{tito2021document} are selected to evaluate model's performance in text-heavy scenarios. 
For baselines, we report our reproduced results rather than reported ones for fair comparison.

\begin{table}[h]
\caption{
LVLM performance with and w/o synthesized detail caption data. 
``Self” indicates detail caption generated by LLaVA-1.5-7B/13B directly, while ``Syn” means data constructed through our five-stage pipeline by the corresponding model. 
$\text{DC}_{100}$ and $\text{DC}_{4870}$ indicates the proposed detail caption evaluation dataset. 
Bold number indicates the best result under the same setting. 
}
\resizebox{\textwidth}{13.5mm}{
\begin{tabular}{l | c c c c c c c c c c c c | c}
\toprule
DC Data & $\text{MME}_{\text{p}}$ & $\text{MME}_{\text{c}}$ & $\text{MMMU}_{\text{v}}$ & MMStar & GQA & VisWiz & POPE & $\text{RefCOCO}_{\text{g}}$ & $\text{OCR}_{\text{Bench}}$ & $\text{VQA}_{\text{Doc}}$ & $\text{DC}_{\text{100}}$ & $\text{DC}_{\text{4870}}$ & Win
\\
\midrule
\rowcolor{gray!20}
\multicolumn{14}{c}{\textit{LLaVA-1.5-7B}} \\
Base & $1487.1$ & $\bm{260.4}$ & $34.6$ & $33.33$ & $62.86$ & $53.70$ & $86.22$ & $72.16$ & $316$ & $28.75$ & $51.26$ & $51.45$ & $-$ \\
+ Self 25k & $1499.3$ & $258.6$ & $36.3$ & $33.40$ & $62.64$ & $54.90$ & $86.84$ & $\bm{72.75}$ & $316$ & $29.26$ & $51.49$ & $51.83$ & $10/12$ \\
+ Syn 25k & $\bm{1523.2}$ & $257.1$ & $\bm{37.3}$ & $\bm{33.53}$ & $\bm{62.86}$ & $\bm{56.88}$ & $\bm{87.08}$ & $72.61$ & $\bm{321}$ & $\bm{30.01}$ & $\bm{51.91}$ & $\bm{52.65}$ & $\bm{11/12}$ \\
\rowcolor{gray!20}
\multicolumn{14}{c}{\textit{LLaVA-1.5-13B}} \\
Base & $1553.4$ & $267.1$ & $34.3$ & $34.80$ & $63.36$ & $58.35$ & $85.90$ & $74.51$ & $331$ & $30.60$ & $51.96$ & $52.05$ & $-$ \\
+ Self 50k & $1543.4$ & $286.8$ & $34.3$ & $\bm{35.40}$ & $63.53$ & $\bm{59.08}$ & $86.17$ & $74.63$ & $331$ & $30.66$ & $\bm{52.62}$ & $52.85$ & $11/12$ \\
+ Syn 50k & $\bm{1564.0}$ & $\bm{286.8}$ & $\bm{34.3}$ & $35.27$ & $\bm{63.56}$ & $58.56$ & $\bm{86.28}$ & $\bm{74.65}$ & $\bm{333}$ & $\bm{30.79}$ & $52.56$ & $\bm{52.91}$ & $\bm{12/12}$ \\
\bottomrule
\end{tabular}
}
\centering
\vspace{0.5mm}
\label{tbl: detailcaps_training}
\vspace{-3mm}
\end{table}

\paragraph{Synthesized detail caption data improves LVLM's overall understanding performance effectively.}
As shown in Table~\ref{tbl: detailcaps_training}, even if we only add a little fraction of synthesized high-quality detail caption data in the SFT stage (25k for 7B model and 50k for 13B model), performance improvement is witnessed across a series of visual understanding benchmarks, demonstrating the effectiveness enhancing LVLM's overall understanding capabilities with synthesized detail caption data. 

\paragraph{Directly generated detail caption data also improves LVLM's overall understanding performance.}
As shown in Table~\ref{tbl: detailcaps_training}, training with detail caption data generated directly also leads to an overall performance improvement. 
Although the improvement is eclipsed by synthesized detail caption data, this observation validates the importance of using detail caption data for model training, even if the data is generated by the model itself directly.

\paragraph{Model's benchmark scores correlate to their detail caption task performance positively.}
We observe a positive correlation between LVLMs' benchmark scores (win rates) and their performance in detail caption tasks. 
This observation validates the importance of detail image captioning task and the feasibility of enhancing LVLM's overall visual understanding abilities by improving its detail caption ability with synthesized high-quality caption data.

\section{Limitations and Future Work}
\label{sec: limitations}

The proposed detail image caption evaluation metric achieves outstanding consistency with human evaluation in the curated benchmarks. 
However, we point out that although two powerful expert are adopted for evaluation dataset construction, these captions may not be perfect. 
Human refining and more reference captions will be incorporated into the detail caption benchmark in out future work. 
For the data construction pipeline, we observe a diminishing effect when the backbone LVLM becomes stronger. 
For example, LVLMs like LLaVA-NEXT and Mini-Gemini uses GPT-4V-annotated detail caption data for training, and therefore the advantage of the proposed pipeline may suffer from incompatible capabilities of other vision and language tools used in the pipeline. 
We will seek to further improve LVLM's detail captioning abilities with more powerful and scalable vision and language suites in our future work.

\section{Conclusions}
\label{sec: conclusions}

In this work, we analyze the shortcomings of existing image caption benchmarks for LVLM evaluation, and curate high-quality expert-annotated evaluation dataset for detail caption evaluation. 
We also propose a novel detail image caption metric CAPTURE, which extracts visual elements from detail captions, and match them through three stages to produce evaluation results.
Experiments show that CAPTURE metric achieves the highest consistency with expert judgements, and ablation studies demonstrate the effectiveness of the stop words filtering module, three-stage matching module and the default ratio of different type of visual elements.
Guided by the proposed detail caption evaluation methods, we further seek to unleash LVLM's detail image captioning ability with a divide-and-conquer caption construction pipeline powered by open-source vision and language tools. 
Experiments show that the proposed pipeline improves LVLM-annotated detail caption data quality significantly, and the data quality can be further improved via self-looping. 
Ablation studies validate the effectiveness of the pipeline design. 


\newpage
\bibliographystyle{plainnat}
\bibliography{main}

\appendix
\newpage
\section{Prompt Templates for Detail Caption Benchmark Curation}
\label{sec: app.prompt}
\begin{figure}[h]
\centering
\includegraphics[scale=0.385]{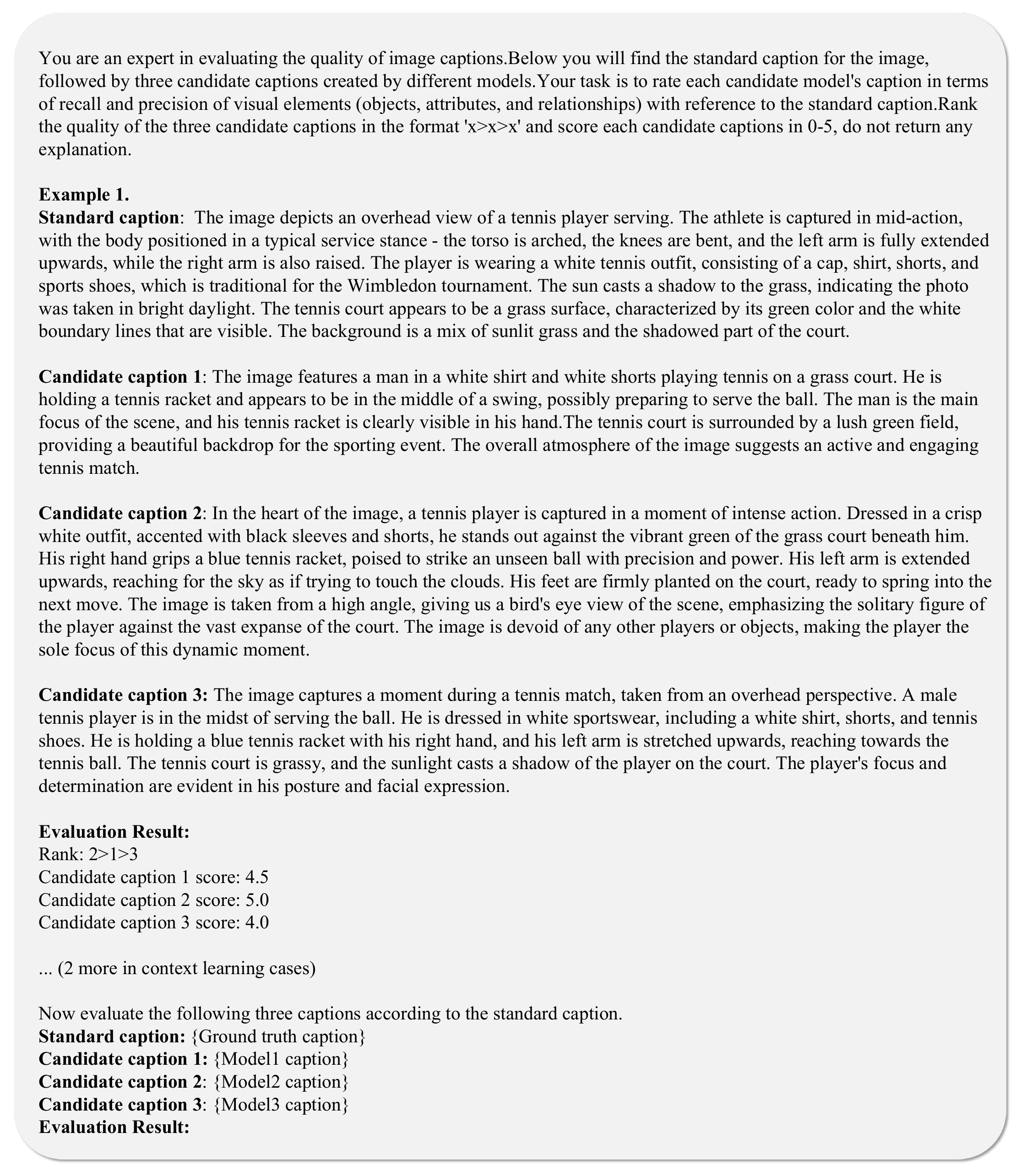}
\caption{
Prompts used to construct the GPT4-Evaluation Score of DetailCaps-4870 dataset. We prompt GPT4 to generate the relative ranking and the absolute score of candidate captions, through three in context learning cases written by human.
}\label{fig: data_prompt}
\end{figure}

\paragraph{Prompt of GPT4-Evaluation scores generation.}
In order to verify the effect of the proposed CAPTURE metric on a larger evaluation set, we use GPT4\cite{achiam2023gpt} instead of humans for evaluation. To better align with human preferences, we manually construct three in context learning cases as shown in Figure \ref{fig: data_prompt}. In each case, a standard caption and three candidate captions are given, and the corresponding human evaluation results are listed as references, including the relative ranking and the absolute scores. Finally, the current ground truth and candidate captions to be evaluated are given in the same format, prompting GPT4 to output the corresponding evaluation results. And we select the output captions of LLaVA-1.5\cite{liu2024visual}, CogVLM\cite{Wang2023CogVLMVE} and ShareCaptioner\cite{chen2023sharegpt4v} as three candidates for evaluation.

\begin{figure}[h]
\centering
\includegraphics[scale=0.285]{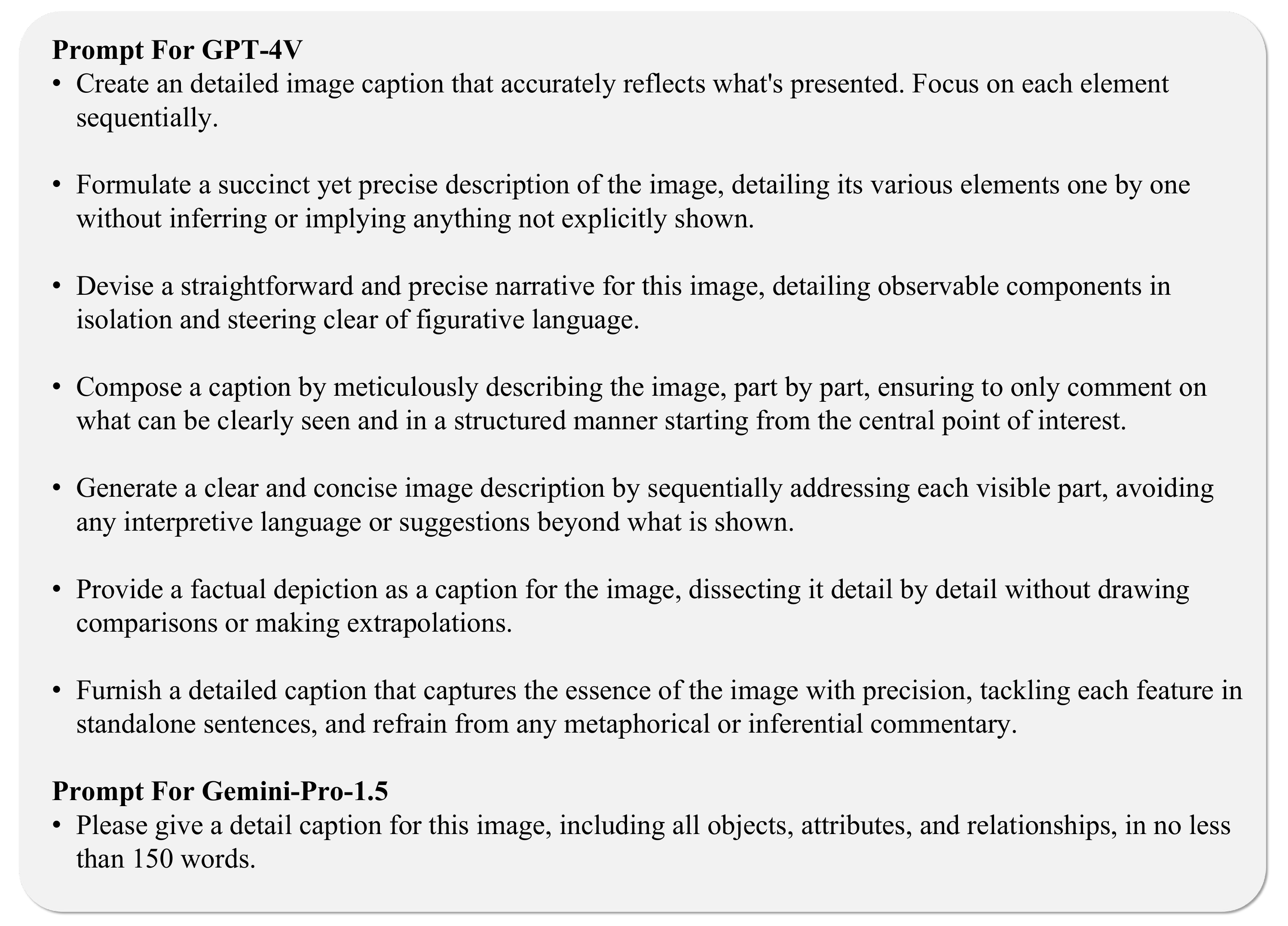}
\caption{
Prompts used to generate detail caption by GPT-4V and Gemini-Pro-1.5.
}\label{fig: detail_prompt}
\end{figure}

\paragraph{Prompt of detail caption generation.}
In the process of generating detail captions, we use multiple different prompts for GPT-4V\cite{gpt4v} to obtain diverse captions as shown in Figure \ref{fig: detail_prompt}. For Gemini-Pro-1.5\cite{reid2024gemini}, we found that the model is more likely to output short captions when the prompt does not indicate the expected output length. Based on this, we only use a single prompt with a word limit for generation.
\label{sec: app.detail_prompt}

\section{Implementation Details for CAPTURE Metric}
\label{sec: app.capture}
\paragraph{Core information extraction. }
\label{sec: 3.2.1}
Core information extraction module aims to extract objects, attributes and relations from a given caption for following matching modules. 
We adopt a SOTA text scene graph parser: Factual parser~\cite{li2023factual} as the backbone model. 
Factual parser is a T5-base model trained on human-annotated scene graph parsing dataset. 
It takes as input a short caption paragraph, and produce the objects, attributes and relations appearing in the caption. 
Since Factual parser is trained on short caption parsing dataset, its performance deteriorates severely when given detail image captions. 
To solve this problem, we first use NLTK toolkit~\cite{bird2006nltk} to cut detail image caption into short paragraphs, and apply Factual parser to each paragraph to obtain a list of parsing results. 
The parsing results are then merged into a larger scene graph based on the following rules: 
(1) all nouns and adjectives are lemmatized with Wordnet~\cite{miller1995wordnet};
(2) duplicated objects are merged as one element, so are corresponding attributes; 
(3) attributes describing two or more merged objects are deduplicated; 
(4) duplicated relations are merged as one element; 
In this way, we obtain a large scene graph for each caption with duplicated elements removed. 
The scene graph is then used to compute the final matching score. 

\paragraph{Stop words filtering. }
Although yielding relatively satisfying parsing results, Factual parser struggles to discriminate concrete nouns from abstract ones, which are not expected to participate in the following matching process. 
For example, in caption "Two white sheep are enjoying the moment", "sheep" refers to a perceptible element in the image, while "moment" has no tangible meaning. 
We filter out abstract nouns via a stop word list: once an object in parsing results appear to be in the stop word list, the word itself will not participate in the object elements matching process. 

To construct such the stop word list, we first apply LLaMA2-13b-chat~\cite{touvron2023llama} and Factual parser to ShareGPT4V-102k dataset for nouns extraction, respectively. 
We observe that LLaMA may omit a proportion of objects appearing in the caption, but the extracted concrete nouns demonstrate impressive precision. 
Based on this obeservation, we curate words recalled by Factual parser but omitted by LLaMA, and compute the frequency of these words. 
Human experts are tasked to judge whether words with the highest frequency are concrete nous or abstract ones. 
Finally, 500 abstract nouns with the highest frequency are curated to be the stop word list. 

It is also worth noticing that although yielding relatively satisfying parsing results, Factual parser struggles when dealing with cross-sentence pronoun reference. 
When given ambiguous pronoun references, Factual parser may generate objects which are not contained in the caption. 
To tackle this problem, we further check the parsed objects' appearance in the caption, and filter out unmatched objects as well as its corresponding attributes and relations. 

\paragraph{Core information matching. }
After obtaining and filtering core information from both ground truth detail caption and candidate one, the extracted elements are matched to produce final evaluation result. 
Intuitively, identical object, attribute or relation elements are matched. 
However, due to the diverse writing style of LVLMs, the same element can be expressed in various ways, and exact matching strategy fail to handle such cases. 
To solve this problem, we add a synonym matching module after exact matching to match elements with similar meanings. 
We employ Wordnet to get the synonym set of both the candidate element and ground truth one, and match them if their synonym sets overlaps. 
Matched candidate objects, attributes and relations are formulated as: 
\begin{equation}
\begin{split}
& cand_{type}^{match} = cand_{type}^{ex} \bigcup cand_{type}^{syn},\\
\end{split}
\end{equation}
where ${type} \in \{obj, attr, rel\}$. $cand_{type}^{ex}$ and $cand_{type}^{syn}$ stand for exactly matched and synonym matched candidate phrases, respectively. 
Matched ground truth elements are formulated in the same way as $gt_{obj}^{match}$, $gt_{attr}^{match}$ and $gt_{rel}^{match}$. 

Exact matching and synonym matching strategies tackle most of the matched cases, but still fail to cover all core information extracted from captions in various writing styles. 
To this end, we propose a soft matching strategy, which takes Sentence BERT~\cite{reimers-gurevych-2019-sentence} model to encode remaining object, attribute or relation phrases and compute a matching score in $[0, 1)$ for remaining unmatched phrases. 
Let $cand_{type}^{rm}$ be unmatched candidate phrases and $gt_{type}^{rm}$ be ground truth ones, their similarity matrix $S_{type}^{rm} \in \mathbf{R}^{|cand_{type}^{rm}| \times |gt_{type}^{rm}|}$ is calculated as:
\begin{equation}
S_{type}^{rm} = \phi(cand_{type}^{rm}) \times \phi(gt_{type}^{rm})^{T}, 
\end{equation}
where $\phi(\cdot)$ denotes Sentence BERT model. 
We further compute the matching score of $cand_{type}^{rm}$ and $gt_{type}^{rm}$ as follows:
\begin{equation}
\begin{split}
& cand\_match_{type}^{rm}[i] = \max_{j=1,2,...,|gt_{type}^{rm}|} S_{type}^{rm}[i, j], \\
& gt\_match_{type}^{rm}[j] = \max_{i=1,2,...,|cand_{type}^{rm}|} S_{type}^{rm}[i, j]. \\
\end{split}
\end{equation}
$cand\_match_{type}^{rm}$ and $gt\_match_{type}^{rm}$ are then used as a complementary to exact matched and synonym matched relations. 

After obtaining matching results, we compute the precision and recall for each type of core information. 
The precision and recall are computed as:
\begin{equation}
\begin{split}
& precision_{type} = \frac{|cand_{type}^{match}|}{|cand_{type}|}, \\
& recall_{type} = \frac{|gt_{type}^{match}|}{|gt_{type}|}. \\
\end{split}
\end{equation}
Attribute precision and recall are computed in the same way. 
As for relation elements, candidate matching score and ground truth matching score are counted separately due to the introduction of feature matching: 
\begin{equation}
\begin{split}
& precision_{type} = \frac{|cand_{type}^{match}|+\frac{\sum{cand\_match_{type}^{rm}}}{|cand\_match_{type}^{rm}|}}{|cand_{type}|}, \\
& recall_{type} = \frac{|gt_{type}^{match}|+\frac{\sum{gt\_match_{type}^{rm}}}{|gt\_match_{type}^{rm}|}}{|gt_{type}|}. \\
\end{split}
\end{equation}
Finally, CAPTURE metric takes the precision and recall of all three types of core information into consideration, and produce the final evaluation result as:
\begin{equation}
CAPTURE = \frac{\alpha F1_{obj} + \beta F1_{attr} + \gamma F1_{rel}}{\alpha + \beta + \gamma}, 
\end{equation}
where $\alpha$, $\beta$ and $\gamma$ are scale factors, and $F1_{type}=\frac{precision_{type} \cdot recall_{type}}{precision_{type} + recall_{type}}$ stands for the F1 score of each type of core information. 

\section{Visualized Examples for Improved Detail Caption Construction}
\label{sec: app.visualization}

\paragraph{Cases of detail caption construction.}
\begin{figure}[h]
\centering
\includegraphics[scale=0.216]{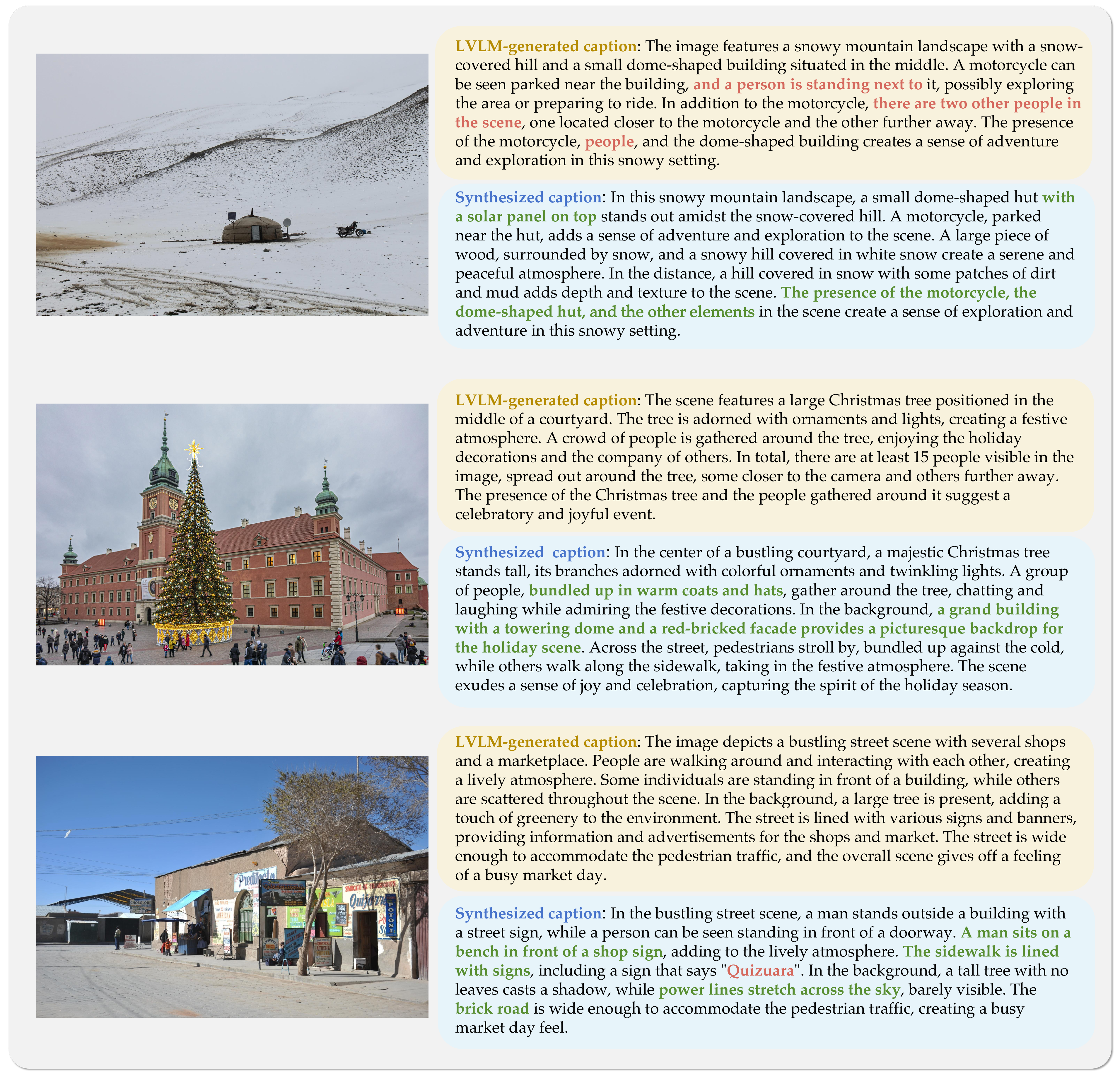}

\caption{
Comparison of the original LVLM-generated caption and the synthesized caption after detail caption construction. The red annotations represent description errors, and the green annotations in the synthesized captions represent the correct descriptions compared to the LVLM-generated ones.
}\label{fig: pipeline_case}
\end{figure}

In Figure \ref{fig: pipeline_case}, we show the effectiveness of detail caption construction in Section \ref{sec: detail_cap_con} with three visualized cases. 
In the first case, highlighted in red, the LVLM-generated caption incorrectly mentions that there are people in the image, while the caption produced by our pipeline removes the relevant description correctly. 
In the following two cases, the synthesized captions complement model-generated captions with additional visual information highlighted in green, resulting into higher-quality detail image caption.

\clearpage

\end{document}